\documentclass{elsarticle}

\usepackage{lineno,hyperref}

\usepackage{subcaption,siunitx,booktabs}

\usepackage{multirow}
\usepackage{booktabs}
\usepackage{scrextend}
\usepackage{tablefootnote}

\journal{Expert Systems With Applications}

\biboptions{authoryear}

\begin{document}
	
	\begin{frontmatter}

		\title{Approaching sales forecasting using recurrent neural networks and transformers}
		
		\author[UV]{Iván Vallés-Pérez}
		\author[UV]{Emilio Soria-Olivas}
		\author[UV]{Marcelino Martínez-Sober}
		\author[UV]{Antonio J. Serrano-López\footnote{Corresponding author. Email antonio.j.serrano@uv.es}}
		\author[UV]{Juan Gómez-Sanchís}
		\author[UV]{Fernando Mateo}

		\address[UV]{Escola Tècnica Superior d\textsc{\char13}Enginyeria, University of Valencia, Avenida de la Universitat s/n 46100 Burjassot, Valencia, Spain}
		\address{\{ivan.valles,
			emilio.soria,
		    marcelino.martinez, antonio.j.serrano, juan.gomez-sanchis,
		    fernando.mateo
			\}@uv.es}
		\begin{abstract}
			Accurate and fast demand forecast is one of the hot topics in supply chain for enabling the precise execution of the corresponding downstream processes (inbound and outbound planning, inventory placement, network planning, etc). We develop three alternatives to tackle the problem of forecasting the customer sales at day/store/item level using deep learning techniques and the \textit{Corporación Favorita} data set, published as part of a \textit{Kaggle} competition. Our empirical results show how good performance can be achieved by using a simple \textit{sequence to sequence} architecture with minimal data preprocessing effort. Additionally, we describe a training trick for making the model more time independent and hence improving generalization over time. The proposed solution achieves a RMSLE of around 0.54, which is competitive with other more specific solutions to the problem proposed in the \textit{Kaggle} competition
		\end{abstract}
		
		
	\end{frontmatter}

	\section{Introduction}
	We have observed how the retail industry economic activity is moving online. In the last years, e-commerce companies are gaining more and more adepts every day. \textit{E-Marketer} (\cite{emarketer2019, emarketer2020}) reported consistent year on year growths in number of sales of more than 13\% in the last 5 years in the US. The percentage of total sales in the US owned by e-commerce companies increased from 8,9\% to 14,5\% in 2020 (\cite{emarketer2019, emarketer2020}).
	
	The continuously increasing demand requires the online industries to constantly improve their supply chain systems. This process entails multiple challenges such as: optimal inventory placement (\cite{graves2008}), accurate network expansion (\cite{hossein2017}), precise inbound and outbound planning (\cite{kaipia2009}), etc. One of the most important wires that enables all those improvements is the ability to accurately forecast the customer demand for different products, locations and times (\cite{forslund2007}).  
	
	This paper proposes several alternatives to solve the demand forecast problem using deep learning techniques (\cite{Goodfellow2016}). The generalization power of these algorithms enables solving the problem using a single model for all the different locations and products time series, while other algorithms like the \textit{ARIMA} family of models (\cite{Hyndman2018}) only can tackle one product-location time series at a time. 
	
	Two approaches are described in this work: a \textit{sequence-to-sequence} (\textit{seq2seq} hereafter) architecture with product and location conditioning and an adapted \textit{transformer} architecture for time series forecasting. The code used in this work has been published in GitHub: \url{https://github.com/ivallesp/cFavorita}.
	
	 Section \ref{sec:data} describes the data set used in this study and the objective to be forecasted. The previous work done in the field of demand forecast and with the \textit{Corporación Favorita} data is described in section \ref{sec:prevwork}. Section \ref{sec:methods} dives deep into the different forecasting methodologies used, together with the different tricks of each implementation. 
	Finally, sections \ref{sec:results} and \ref{sec:conclusions} detail the results obtained and summarize the conclusions of this work.

	\subsection{Data} \label{sec:data}
	The \textit{Corporación Favorita Grocery Sales} data set  has been used to conduct this study (\cite{corporacionfavoritadataset2018}). \textit{Corporación Favorita}, an Ecuadorian company owner of multiple supermarkets across Latin America, released this data set around 2017 as a \textit{Kaggle} competition to challenge the community to forecast their sales. It contains daily sales records for 4,400 unique items, in 54 different Ecuadorian stores from January 1st 2013 to August 15th 2017. Additional data provided along with the number of sales are described below.
	
	\begin{itemize}
		\item ID variables: date, store number and item number.
		\item Promotions: a binary variable indicating if a given item, in a given store at a given time was on promotion.
		\item Store information: location (city and state) and segment (type and cluster).
		\item Item information: item family, class, and a binary variable indicating if the item is perishable.
		\item Transactions: Number of total sales for each store at each date.
		\item Oil price: price of the oil on each date.
		\item Holidays and events: dates, locations and types of holidays.
	\end{itemize}
	
\begin{figure}
		\centering
		\includegraphics[width=0.48\linewidth]{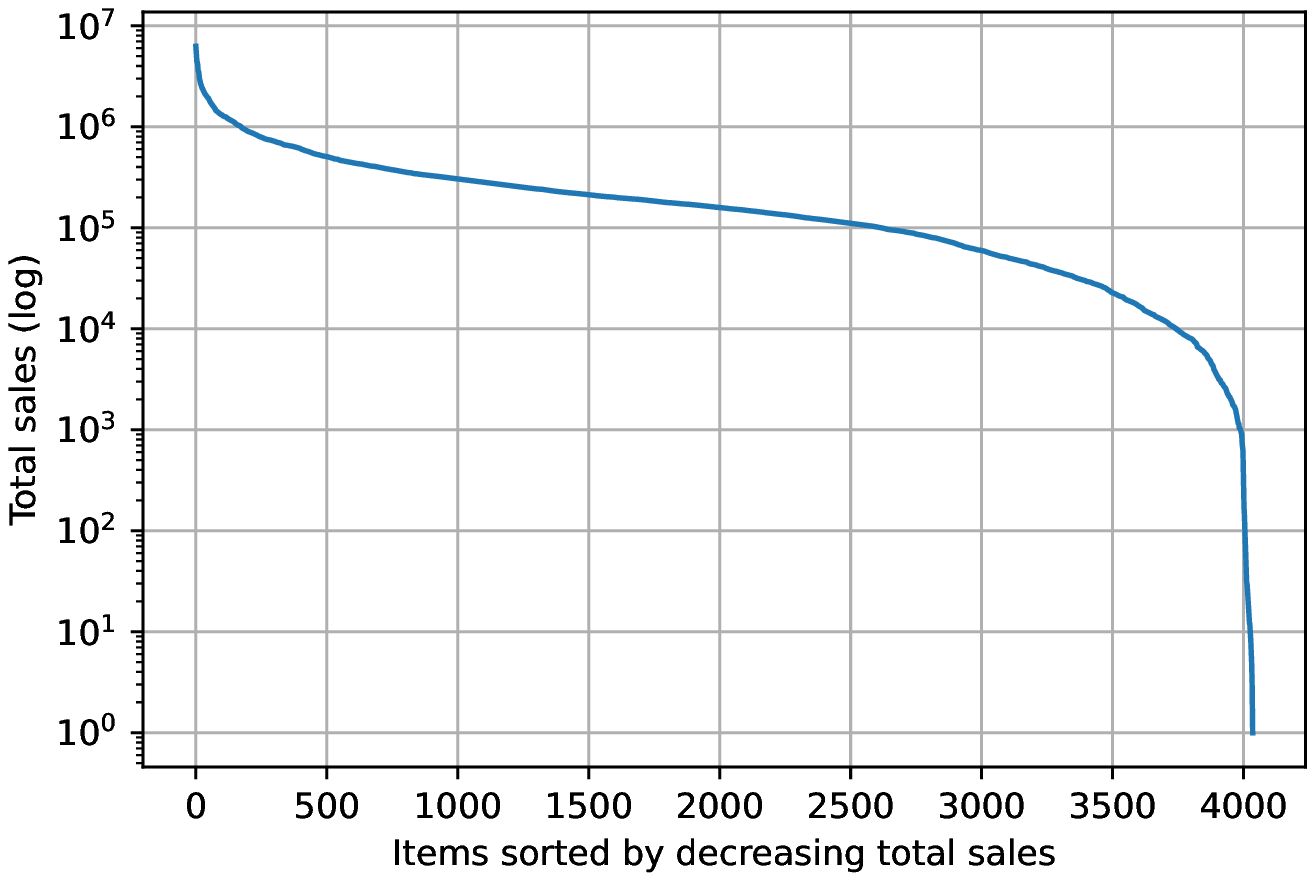}
		\includegraphics[width=0.48\linewidth]{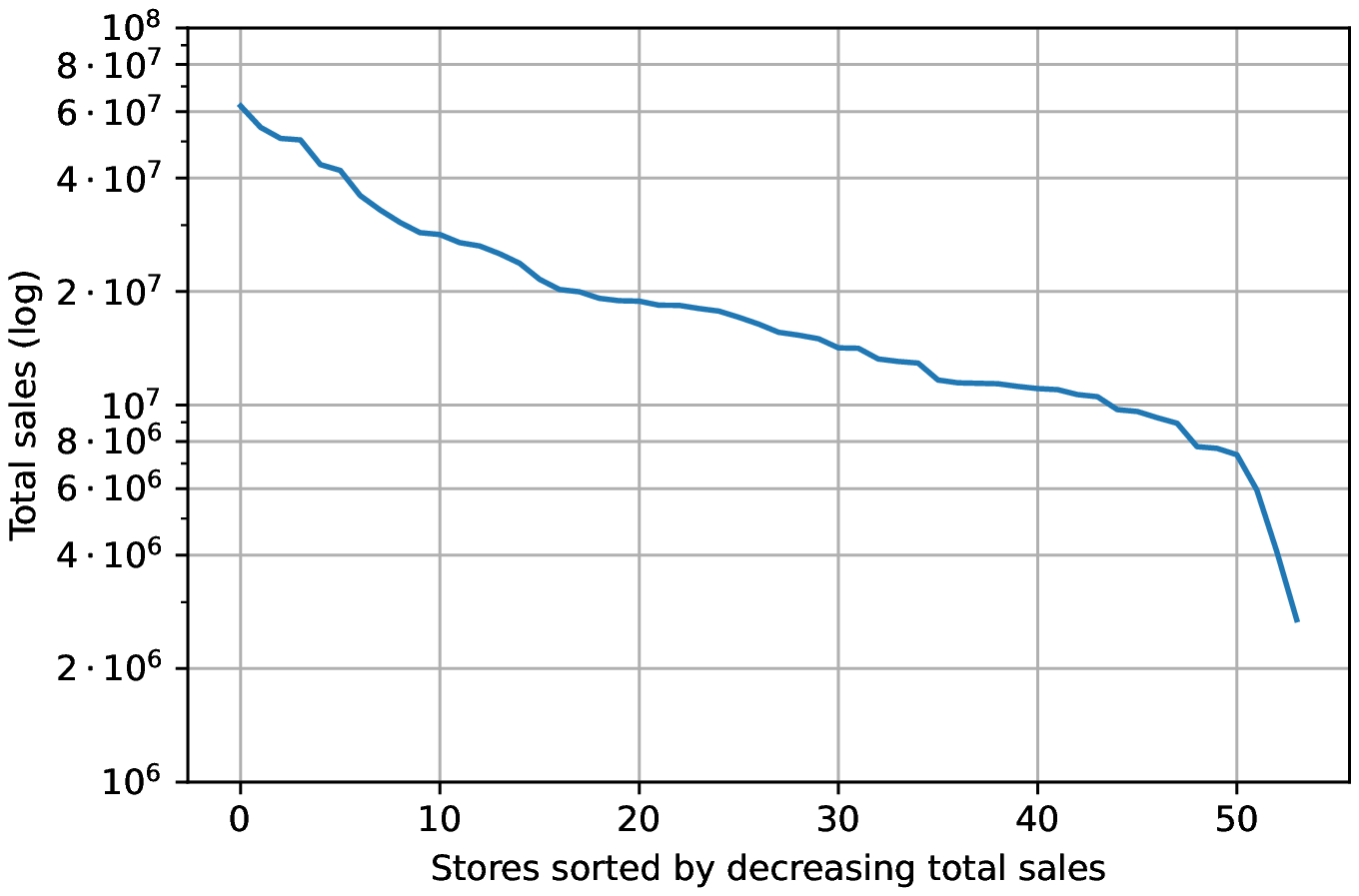}
		\caption{Long tail distributions for items (left) and stores (right). Y axes represent total sells across all the history (around 5 years).}
		\label{fig:tails}
	\end{figure}

	As it is usual in the supply chain organizations, the distribution of sales across products is very uneven (see Figure \ref{fig:tails}-left where the top 10\% of the products bring the 44\% of the sales). The same characteristic is observed in the case of the stores (see Figure \ref{fig:tails}-right where the top 10 stores bring 40\% of the sales). The sales for some products are very sparse: during the test period, the probability of having one or more sells for a given product is 47.6\%, which in practice means that more than half of the days in the time series will have value of zero.
	
\begin{figure}
	 	\centering
	 	\includegraphics[width=1\textwidth]{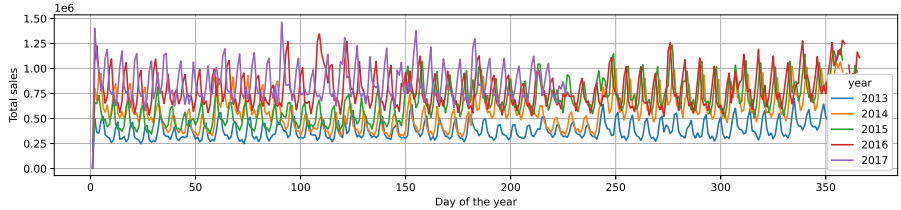}
	 	\caption{Daily total sales for the 5 years included in the data.}
	 	\label{fig:timeseries}
	 \end{figure}

\begin{figure}
		\centering
		\includegraphics[width=0.6\textwidth]{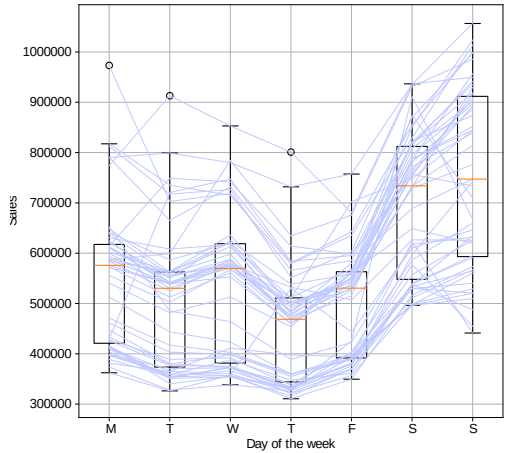}
		\caption{Weekly pattern detail, including the 2014 sales.}
		\label{fig:timeseries_detail}
	\end{figure}

	 As it can be noticed in Figure \ref{fig:timeseries}, the total sales show clear year on year trends as well as a very remarkable weekly seasonality (see Figure \ref{fig:timeseries_detail}). There is also a peak of sales at the end of each year
	 
	The data set does not contain records for items in days when there were zero unit sales. It also lacks information about the available inventory. These two facts together make the forecast effort more complicated, given that when there are zero sales of a given product in a store at a given date, it can be either because there was not available inventory, or because there was inventory but not demand (or both of them at the same time).
	
	Estimating the actual demand of a retailer is not a straightforward task (\cite{Deep2019}). In our case, the quantity being forecasted is the number of sales. That would have an important implication in the demand estimation: the number of sales represent the demand as long as there exists available inventory. Hence, the quantity estimated by the machine learning model will be the demand bounded by the inventory $min(demand(i,s,d), inventory(i,s,d))$ (where $i$, $s$ and $d$ are the inventory, the store and the date). There are techniques to correct the demand in these cases (e.g. \cite{Bell2000}). However as the objective of this study is to build the forecast model, it is out of the scope of this paper to deal with that inconvenience.
	
	\subsection{Previous work}	\label{sec:prevwork}
	We have found several works applying deep learning techniques for demand and sales forecast in the supply chain environment. \cite{Kilimci2019} use several \textit{multi-layer perceptrons} (MLP) to build a unified forecast and compare it with more classical techniques. The main disadvantage of this approach is that it requires heavy feature engineering work, as the MLPs are not suitable to deal with time series. In a supply chain problem, this is not practical given the huge quantity of data normally available. In the work by \cite{Talupula2018}, different deep learning architectures (Convolutional Neural Networks (\textit{CNN}), Long-Short Term Memory (\textit{LSTM}) and Multi Layer Perceptron performance are compared over an outbound demand forecast task, using data from a retailer. Similarly, \cite{Helmini2019} compare a deep learning model based on LSTMs with peephole connections with more classical approaches using tree-based models.  The difference with our approach is that we propose a set of flexible architectures capable to deal with multi-modal data more efficiently, while the models proposed in these works can only deal with time series. Furthermore, we use sequence-to-sequence modeling, which minimizes the error across all the predicted sequence, while the authors of the aforementioned papers use a next-step prediction auto-regressive model, which only optimizes the error of the next time step. 
	
	However, we only found a few reports using this data set for benchmarking. In \cite{kechyn2018}, the authors used the \textit{WaveNet} (\cite{vanderoord2016}) architecture to build an autoregressive forecast. They achieved the 2nd position in the Kaggle competition that published the \textit{Corporación Favorita} data. The authors only show some charts summarizing their results, pointing to Root Mean Squared Weighted Logarithmic Errors (RMSWLE) of around 0.578. However, they do not give many details about the architecture used and the experimental framework. They also do not specify the period of time used to measure the errors. \cite{Steves2018} used multiple classical techniques (historical average, ARIMA/SARIMA, Snaive and exponential smoothing) to forecast the daily sales. These techniques do not consider multiple sale points and products at the same time. They reach a minimum RMSWLE of 0.555. These solutions achieve competitive results. However, they require training one model per item and store (around 238,000 models), which is not a scalable approach for a production environment. The methods proposed in this paper consist of a single model that is used over all the time periods, items and points of sale. 

	Other studies using the same data set have been found: \cite{Wang2020, Shaikhha2020, Schleich2019, Lim2019, Curtin2020, Malik2019, Kuleshov2018, Khamis2020}. Although some of them may be interesting for the supply chain goals, they deviate from our demand/sales forecast goal.
	
	\section{Material and methods} \label{sec:methods}
	The size of the data set chosen (around $4\cdot10^8$ samples) enables the use of deep learning models. Two different neural architectures have been designed: a \textit{seq2seq} model and a \textit{transformer} model.
	
	\subsection{Seq2seq}  
	A \textit{sequence-to-sequence} architecture (abbreviated as \textit{seq2seq}) is a model that is trained to map an input sequence to an output sequence, without any length restriction in both sides (input and output sequences can be of different length) (\cite{sutskever2014}). This architecture contains two main blocks: one encoder and one decoder. The encoder consists of a recurrent neural network which processes the input sequence, one sample at a time, condensing all the relevant input sequence information into a fixed length \textit{context vector}. This vector is usually the last hidden state of the encoder $\mathbf{h_t}$ (\cite{kamath2019}). The decoder, also consisting of a recurrent neural network, generates the output sequence conditioned to the \textit{context vector}. Both modules are trained together to minimize an error term over the output.
	 
	\begin{figure}[h!]
		\centering
		\includegraphics[width=1\linewidth]{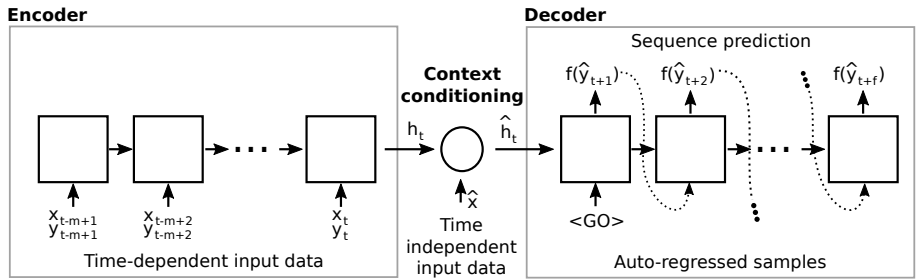}
		\caption{\textit{Seq2seq} architecture diagram. The left box shows the encoder, which takes the historical actual sales $y_i$ as input as well as other exogenous time-dependent features (i.e. oil price, holidays and events, transactions and promotions) denoted by $x_i$, and returns the hidden vector $h_t$ of the last recurrent module as output (aka \textit{context vector}). The middle of the figure shows the context conditioning module, which is our variation proposal over the original sequence to sequence proposal. This module receives the \textit{context vector} $h_t$ from the encoder module and concatenates it with static data $\hat{x}$ producing a conditioned context vector $\hat{h}_t = (h_t, \hat{x})$. Finally, on the right, the decoder module receives as initial state the conditioned context vector and provides the first recurrent cell with a \textit{go symbol} (constant input indicating the input of the decoded sequence). The decoder generates the sequence prediction in an autoregressive way. $f$ is a stack of two fully-connected layers applied to each output of the decoder.}
		\label{fig:s2s}
	\end{figure}
	
   For the purpose of the current study, the original \textit{seq2seq} architecture has been slightly modified to condition the \textit{context vector} to a set of static (non time-dependent) features (see Figure \ref{fig:s2s} for a graphical representation of the architecture). To achieve that, the original context vector $\mathbf{h_t}$ has been concatenated with the static features (item and store embeddings and related features) and then passed into a feed-forward neural network with two hidden layers. The context conditioning module is an important part of the network because it allows the model to adapt the predictions to each product and store. The output of the feed-forward neural network has been used as the initial state of the decoder. The input of the first recurrent cell of the decoder is a special symbol that indicates the model that it is the first step of the output sequence. In this case, the special symbol is a vector containing all zeros. 
   
   The decoder module is a first order auto-regressive model, i.e. the predicted value for time step $t$ is used as input for the prediction of time step $t+1$.

	\subsection{Transformer}
	\textit{Transformer} architectures were firstly published  in 2017 (\cite{vaswani2017}). This architecture removes the need to use recurrent neural networks by implementing attention and self-attention mechanisms (\cite{bahdanau2015}).  Like \textit{seq2seq} architectures, the transformers are able to map an input sequence to an output sequence, with potentially different lengths. Similarly, they also consist of two blocks: the encoder and the decoder.
	
	The attention mechanism can be described as shown in equation \ref{eq:att}. Where $Q$, $K$, $V$ stand for \textit{query}, \textit{key} and \textit{value}, respectively. These three pieces represent an analogy, introduced by \cite{vaswani2017}, of the information retrieval systems where a \textit{query} is used in order to look for the matching \textit{key} (or the most similar one) and retrieving its \textit{value}. The attention mechanism working principle is similar to those systems. There are many possible differentiable similarity functions ($f$) that can be used (\cite{kamath2019}). \cite{vaswani2017} propose the Scaled Dot Product as similarity function, given that it is scaled so that different length sequences can be easily compared together. The Scaled Dot Product is defined in equation \ref{eq:scaled_dot_product},  where $d_K$ represents the length of the key vector $K$. We have adopted this version as it showed to work well in the initial transformer publication.
	
	We applied a slight modification over the original transformer, removing the \textit{softmax} operation of the output and only using the categorical embeddings for the categorical inputs. This is necessary in our case because our task is a regression and not a classification.
	
	\begin{equation}
	C(Q,K,V) = \text{softmax}(f(\mathbf{Q}, \mathbf{K})) \cdot \mathbf{V}
	\label{eq:att}
	\end{equation}
	
	\begin{equation}
	f_{\text{SDP}}(\mathbf{Q}, \mathbf{K}) = \frac{(\mathbf{Q} \cdot \mathbf{K}^T)} {\sqrt {d_K}}
	\label{eq:scaled_dot_product}
	\end{equation}
	
\begin{figure}[h!]
    	\centering
    	\includegraphics[width=0.7\linewidth]{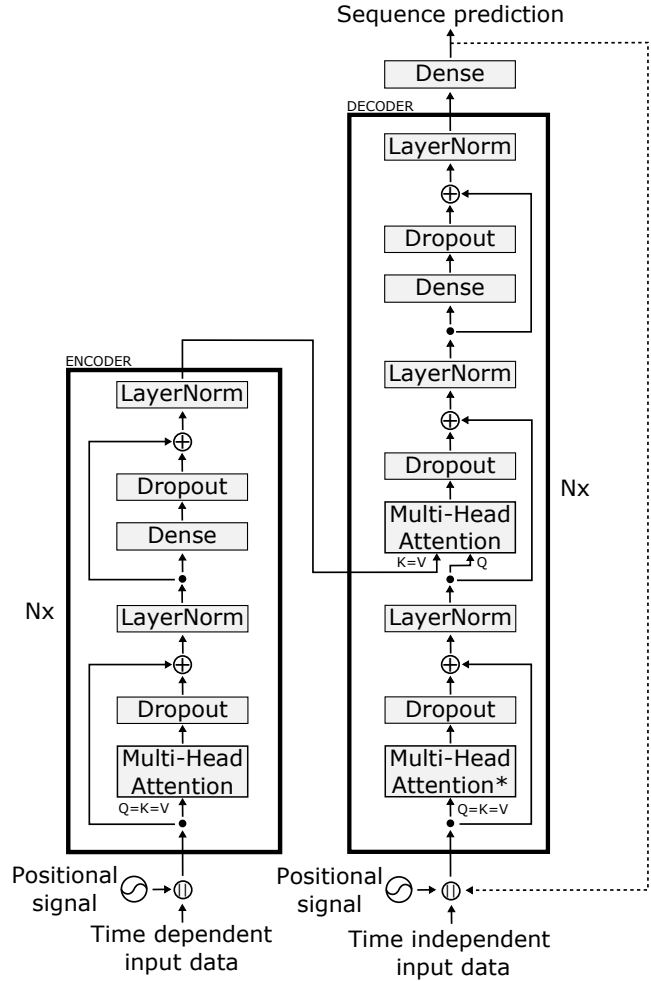}
    	\caption{Transformer architecture used to perform time series forecasting. In the diagram, the $||$ symbol stands for the concatenation operation, and similarly as in \cite{vaswani2017}, $N_X$ represents the number of repetitions of the encoder and decoder blocks.}
    	\label{fig:transformer}
    \end{figure}

	Following the information retrieval analogy and as illustrated in the Figure \ref{fig:transformer}, there are two types of attention being used in this architecture.
	\begin{itemize}
		\item \textit{encoder-encoder attention}: this is a form of self-attention that is used in the encoder module. In it, the \textit{query}, the \textit{key} and the \textit{value} come from the same time series.
		\item \textit{decoder-decoder masked attention}: this is also a form of self-attention with the particularity that the operation is forced to be causal, i.e. it only uses time steps from the past, the future ones are masked out. The \textit{query}, \textit{key} and \textit{value} come from the same time series.
		\item \textit{encoder-decoder attention}: this attention mechanism compares the decoder information with the encoder one, hence it is not self-attention. The \textit{query} comes from the decoder while the \textit{key} and \textit{value} are taken from the encoder output.
	\end{itemize}
	
	To train the \textit{transformer} architecture the \textit{teacher forcing} technique (\cite{williams1989, goyal2016}) is used. It consists in feeding the decoder with the target sequence, right-shifted by one sample so that the training can be done in one single calculation per batch. This technique is commonly used in auto-regressive models to improve the speed of training, and showed good results in the literature (\cite{vaswani2017}). On inference, \textit{teacher forcing} is no longer available (because the future time steps of the time series are unknown) so auto-regression is used to compute the next steps recursively (i.e. the predicted sample is fed back to the input in order to predict the next sample).
    
    \subsection{Random max time step trick} 
    At training time and with the aim of improving generalization over different periods of time, each \textit{minibatch} has been constructed so that the maximum time step (the most recent one)  is drawn randomly from the time line. This trick allows the algorithm to learn a model that generalizes over different periods of time, preventing it to overfit to a single time span.
    
	\section{Experimental setup and results} \label{sec:results}
	
	\subsection{Experimental setup} \label{sec:expsetup}
	The data set provided has intentionally been minimally pre-processed as one of the goals of the current study is to provide a simple and flexible solution to the demand forecast problem. The most important transformation consisted of filling the zero sales records, as the data set was provided without them. The numerical input variables have been normalized by centering and scaling them while the categorical variables have been turned into \textit{one hot encoding} representations. The target variable has been normalized using a logarithmic transformation, as suggested by the authors of the data set in the \textit{Kaggle} competition (\cite{corporacionfavoritadataset2018}). The ID variables corresponding to the store and the item have been used as an input to an embedding lookup layer to give the model the opportunity to learn store or item related information.
	
	The model has been trained using daily data from January 1st 2013 to May 27th 2017, to produce daily forecasts of the next 16 days\footnote{this is not an arbitrary decision, we chose 16 days because that was the requirement in the \textit{Kaggle} competition. That choice would make sense for bi-monthly forecast publications, as it would be applicable for months with 28, 29, 30 and 31 days}. Data from  June 13th 2017 to June 28th 2017 have been used for validation purposes and the next 3 16-days time spans (June 29th to July 14th, July 15th to July 30th and July 31st to August 15th, referred subsequently as period 1, 2 and 3 respectively) have been used to test the performance of the algorithms.
	
	 The \textit{random max time step} has been constrained not to lay before October 29th 2013, to assure that the model has at least 300 days of history to learn from. 
	
	In the \textit{seq2seq} model all the history (from January 1st 2013) has been used as input.  In the \textit{transformer} model, given the quadratic computational complexity dependence on the length of the input sequence, the history had to be shortened to 200 days. In the spirit of fair comparison, an alternative seq2seq version (referred subsequently as \textit{seq2seq trimmed}) has been trained using the 200 most recent time steps in every \textit{minibatch}. To facilitate the interpretation of the results, two baselines have been included: \textit{random} and \textit{average}. The first one consists of measuring the accuracy of a naive prediction built by randomly permuting the target variable. The second one consists of predicting the average of the target variable for all the instances. 

\subsection{Results and discussion} 

	The results have been measured using the \textit{Root Mean Squared Logarithmic Error} (\textit{RMSLE}, defined in equation \ref{eq:rmsle}), \textit{Root Mean Squared Weighted Logarithmic Error} (\textit{RMSWLE}, defined in equation \ref{eq:rmswle}, where the perishable items are given a weight of $1.25$, and $1.0$ to the rest),  and \textit{Mean Absolute Logarithmic Error} (\textit{MALE}, defined in equation \ref{eq:male}). 
	
	\begin{equation} \label{eq:rmsle}	
	RMSLE = \sqrt{ \frac{1}{N} \displaystyle\sum_{i=1}^N  \left(\log(\hat{y}_i + 1) - \log(y_i +1)  \right)^2  }
	\end{equation}
	
	\begin{equation} \label{eq:rmswle}	
	RMSWLE = \sqrt{ \frac{\displaystyle\sum_{i=1}^n w_i \left( \log(\hat{y}_i + 1) - \log(y_i +1)  \right)^2  }{\displaystyle\sum_{i=1}^n w_i}}
	\end{equation}
	
	\begin{equation} \label{eq:male}	
	MALE = \sqrt{ \frac{1}{N} \displaystyle\sum_{i=1}^N  \left|\log(\hat{y}_i + 1) - \log(y_i +1)  \right|  }
	\end{equation}

	In the previous equations, $\hat{y}$ represents the predicted sales,  $y$ represents the actual sales and $N$ is the total number of samples. The logarithmic component of the error metrics was introduced because different products at different shops have arbitrarily different demand levels. The usage of the logarithm normalizes the unit sales distribution and makes the whole problem easier to measure.  We always used the natural logarithm in this study. The \textit{RMSWLE} error metric has been introduced for easier comparison and benchmarking with future studies. 
	
\begin{figure}[h!]
		\centering
		\includegraphics[width=1\linewidth]{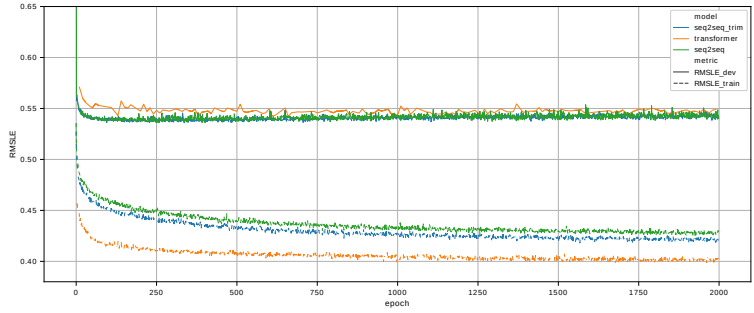}
		\caption{Evolution of the train and validation (dev) error during the process of training, for all the models.}
		\label{fig:performance_evolution}
	\end{figure}

\begin{figure}[h!]
		\centering
		\includegraphics[width=0.48\linewidth]{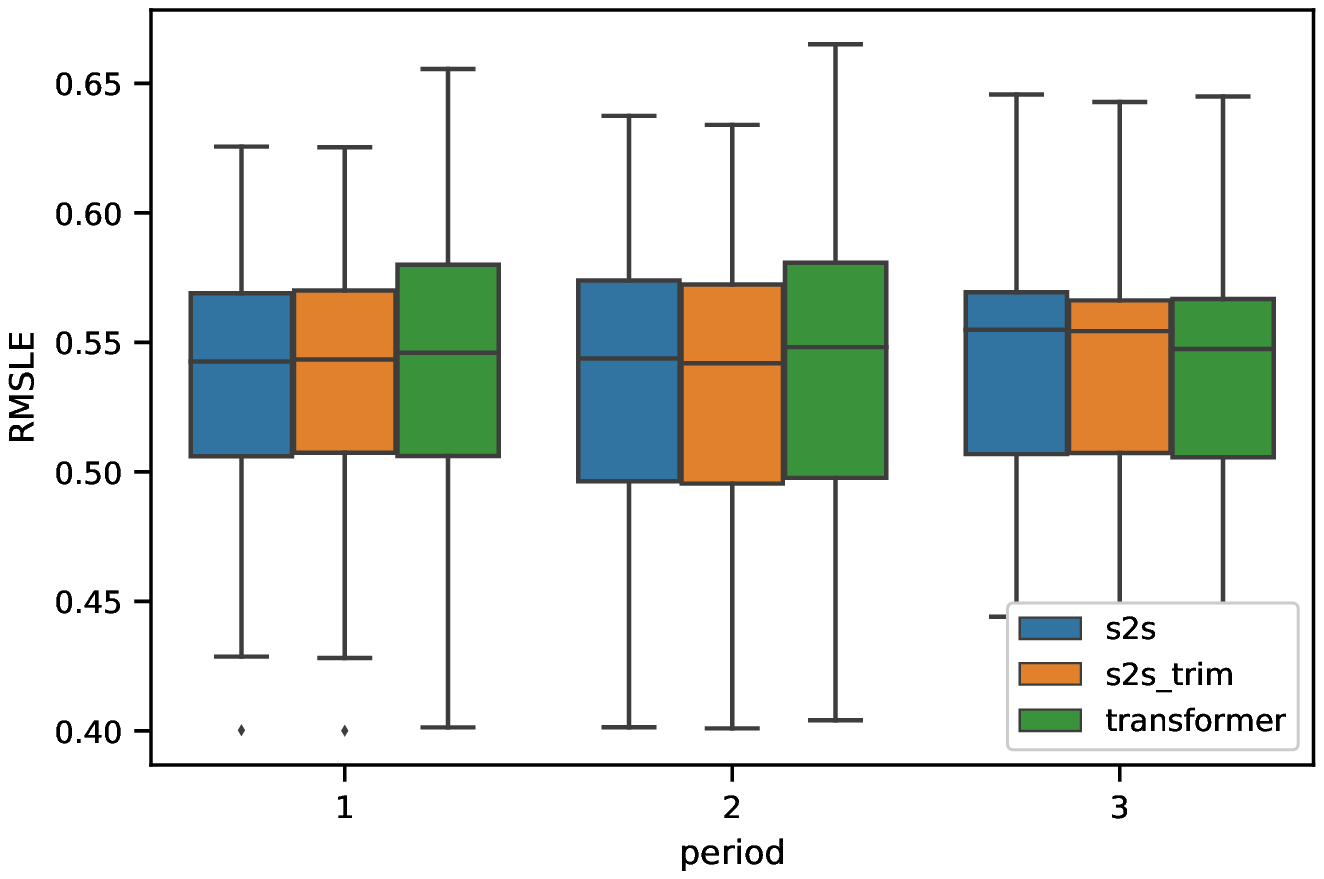}
		\includegraphics[width=0.48\linewidth]{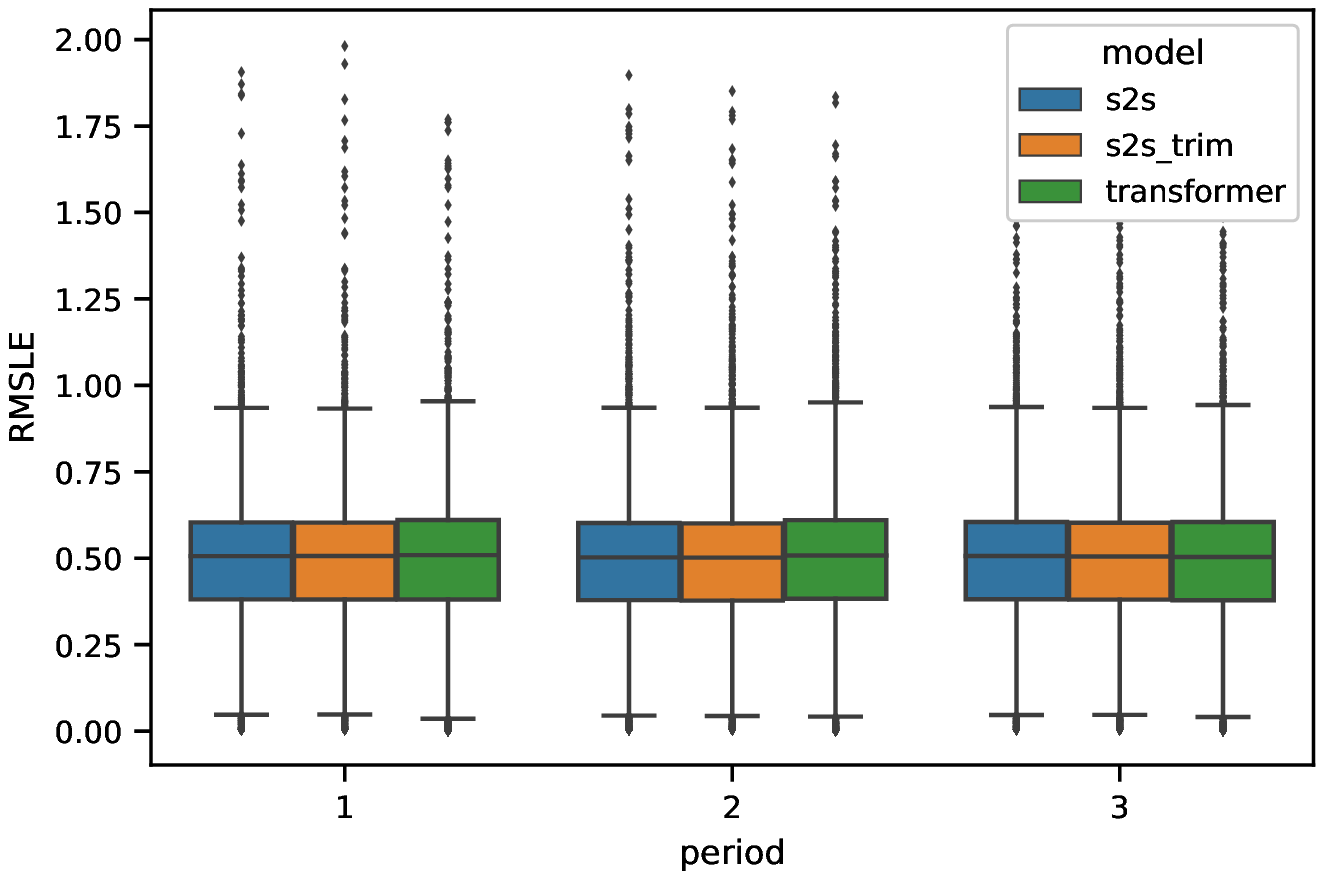}
		\caption{Distribution of the error across stores (left) and accross items (right) for every model  and for the three different test periods used.}
		\label{fig:stores_items_performance}
	\end{figure}

\begin{figure}
		\centering
		\includegraphics[width=1\linewidth]{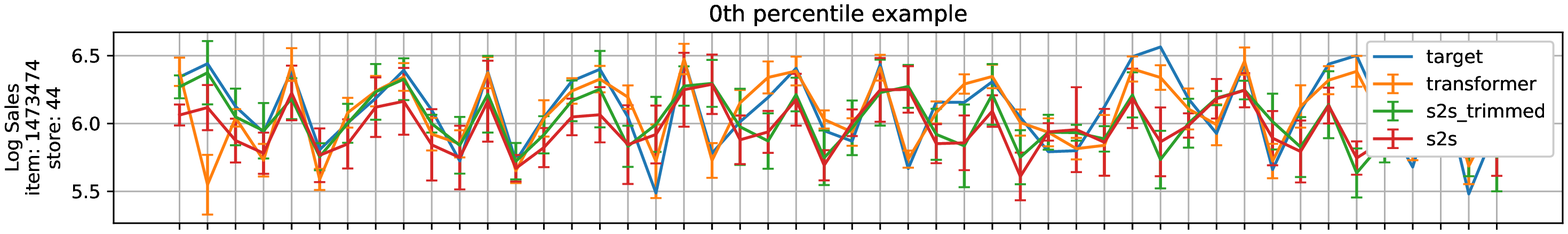}
		\includegraphics[width=1\linewidth]{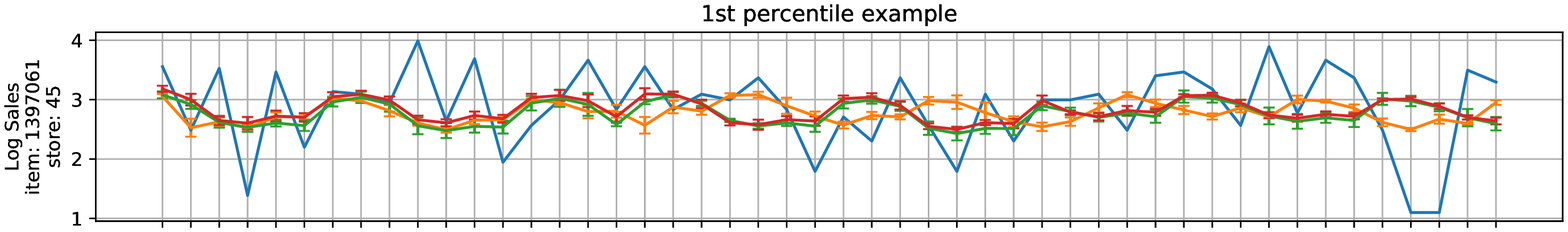}
		\includegraphics[width=1\linewidth]{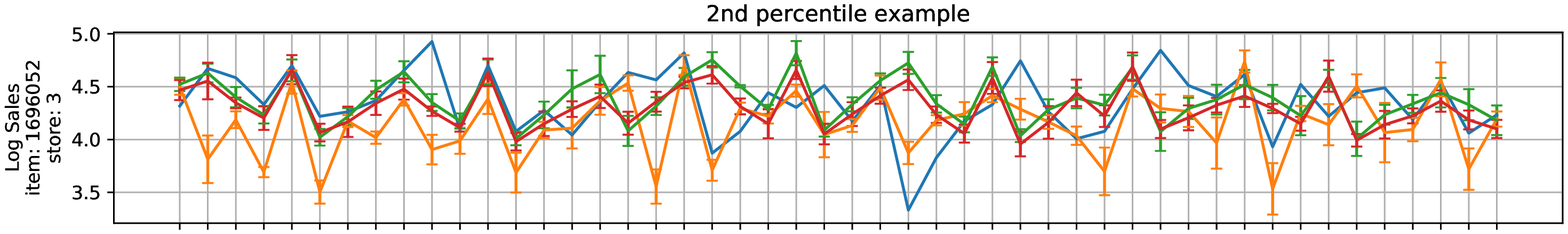}
		\includegraphics[width=1\linewidth]{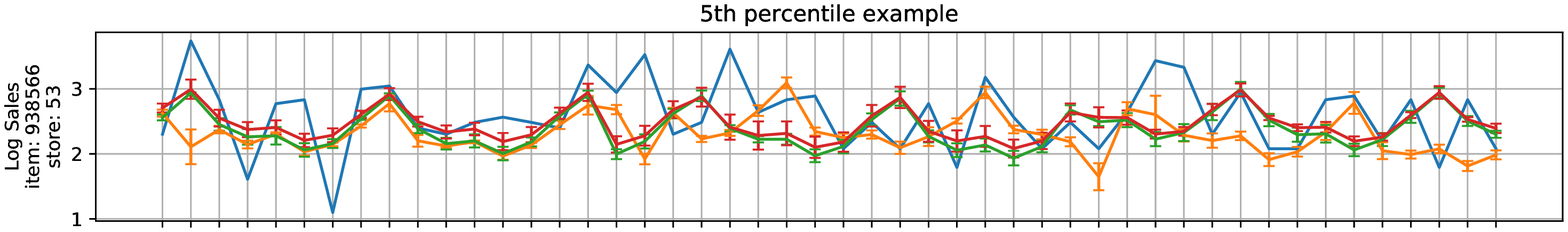}
		\includegraphics[width=1\linewidth]{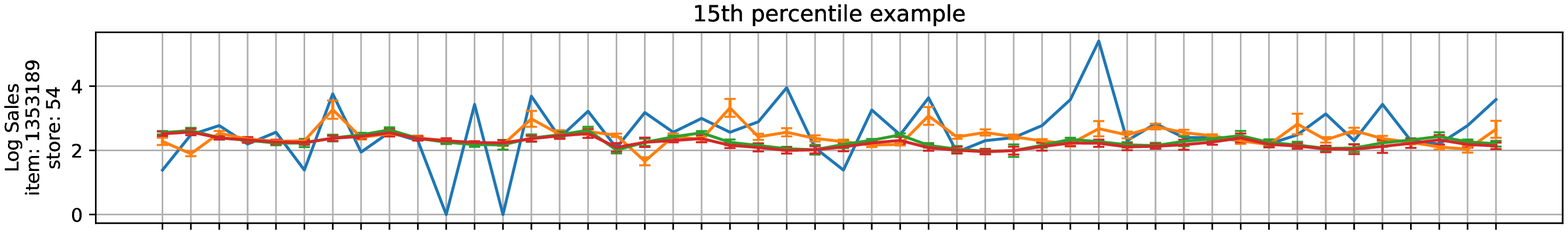}
		\includegraphics[width=1\linewidth]{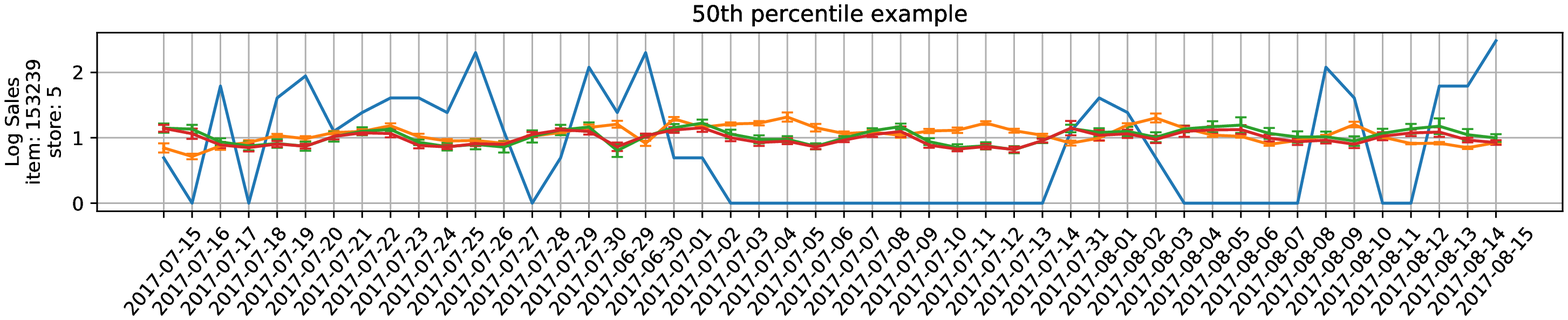}
	\caption{Actual and forecasted sales in log space for six examples of store-item combinations representing the 0th (best prediction), 1st, 2nd, 5th 15th and 50th percentiles of RMSLE (relative to the target variable average) from top to bottom. The three test periods have been concatenated along the X axis. The error bars show the standard deviation across the 5 runs.}
		\label{fig:ts_log}
	\end{figure}

\begin{figure}
	\centering
	\includegraphics[width=1\linewidth]{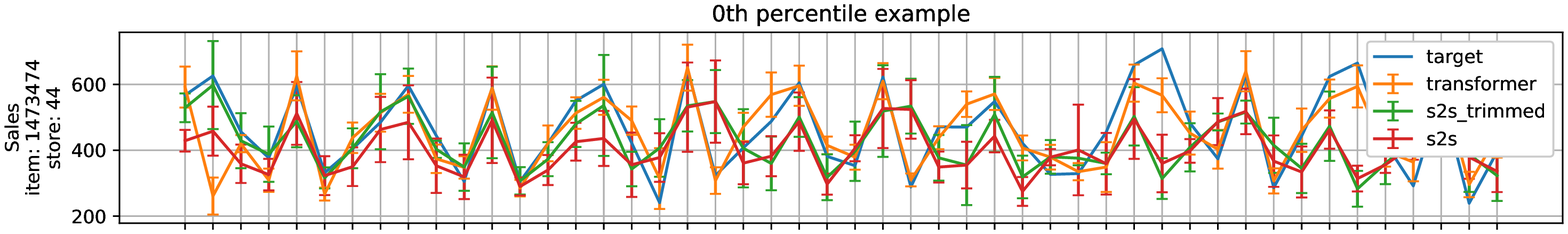}
	\includegraphics[width=1\linewidth]{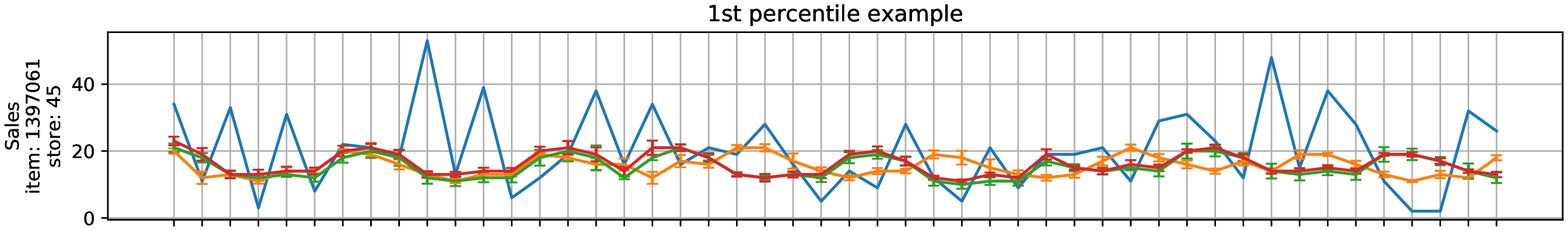}
	\includegraphics[width=1\linewidth]{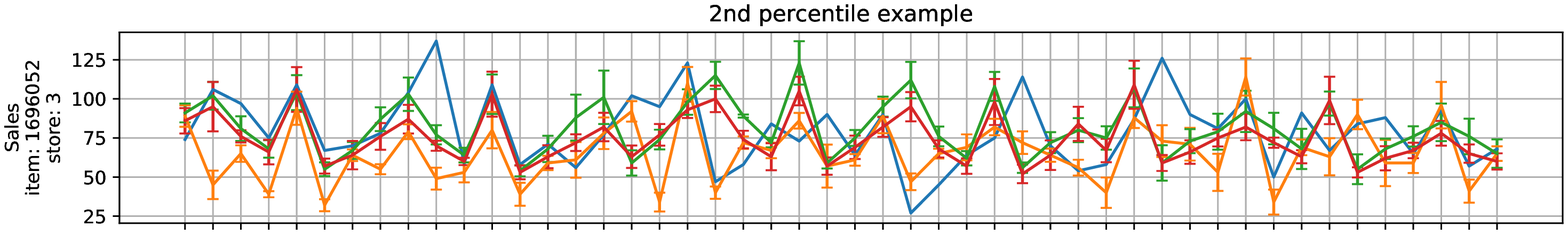}
	\includegraphics[width=1\linewidth]{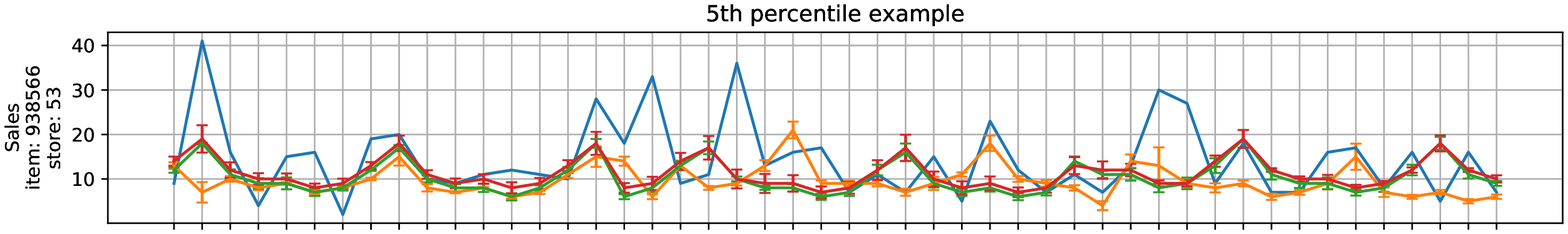}
	\includegraphics[width=1\linewidth]{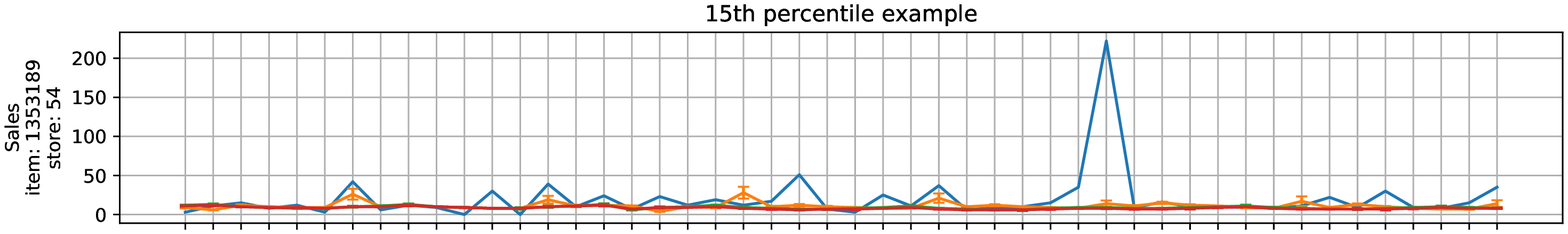}
	\includegraphics[width=1\linewidth]{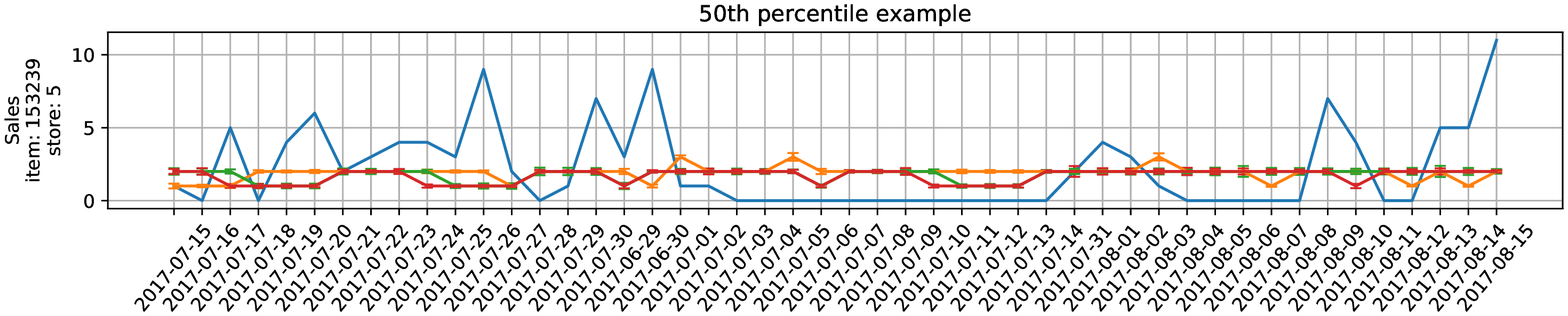}
	\caption{Actual and forecasted sales in linear space for six examples of store-item combinations representing the 0th (best prediction), 1st, 2nd, 5th 15th and 50th percentiles of RMSLE (relative to the target variable average) from top to bottom. The three test periods have been concatenated along the X axis. The error bars show the standard deviation across the 5 runs.}
	\label{fig:ts_lin}
\end{figure}

	Figure \ref{fig:performance_evolution} shows how the errors evolve in every epoch. Figure \ref{fig:stores_items_performance} decomposes the error at store and item level, in order to show in detail how the errors vary along these dimensions. Finally, Figures \ref{fig:ts_log} and \ref{fig:ts_lin} show examples of actual and forecasted time series in log and linear scales, respectively.
	
\begin{figure}
	\centering
	\includegraphics[width=1\linewidth]{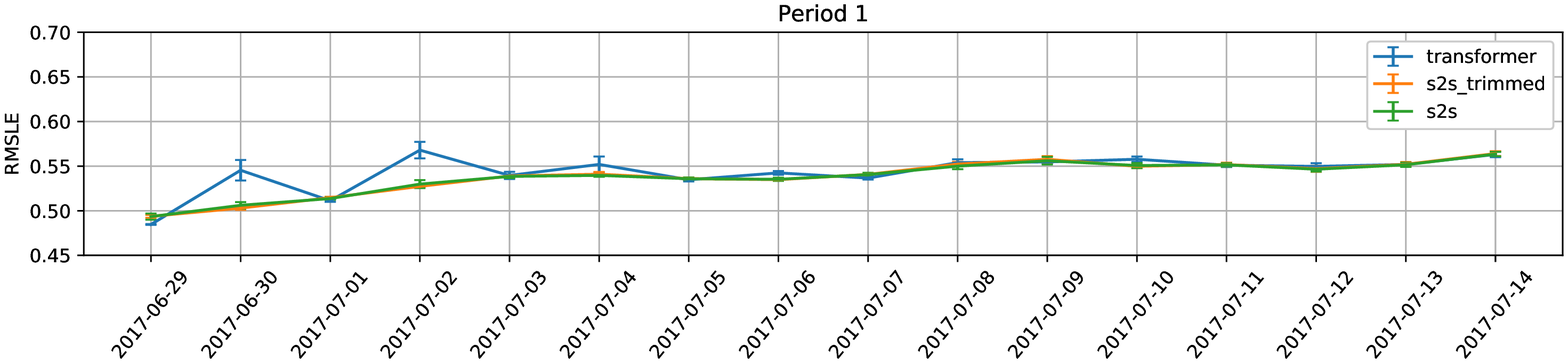}
	\includegraphics[width=1\linewidth]{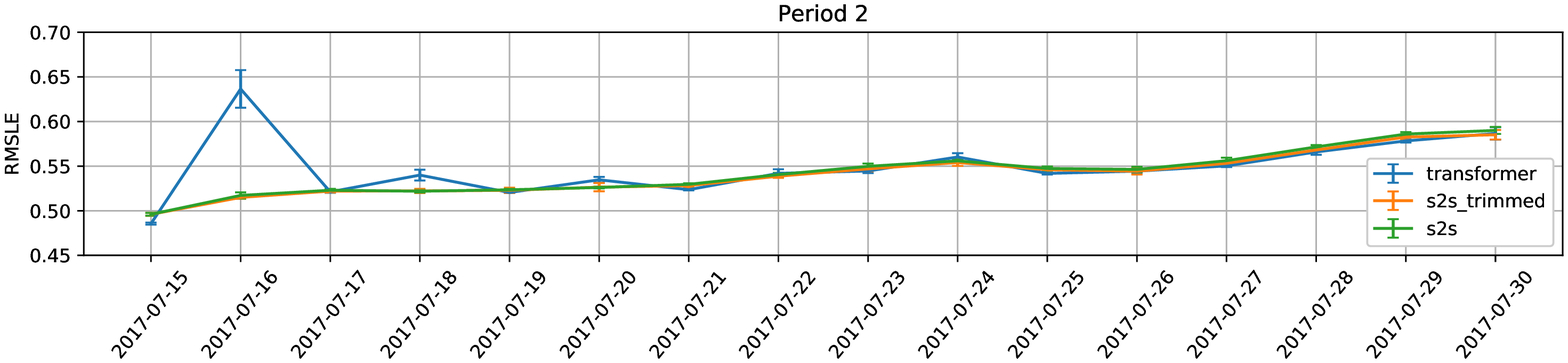}
	\includegraphics[width=1\linewidth]{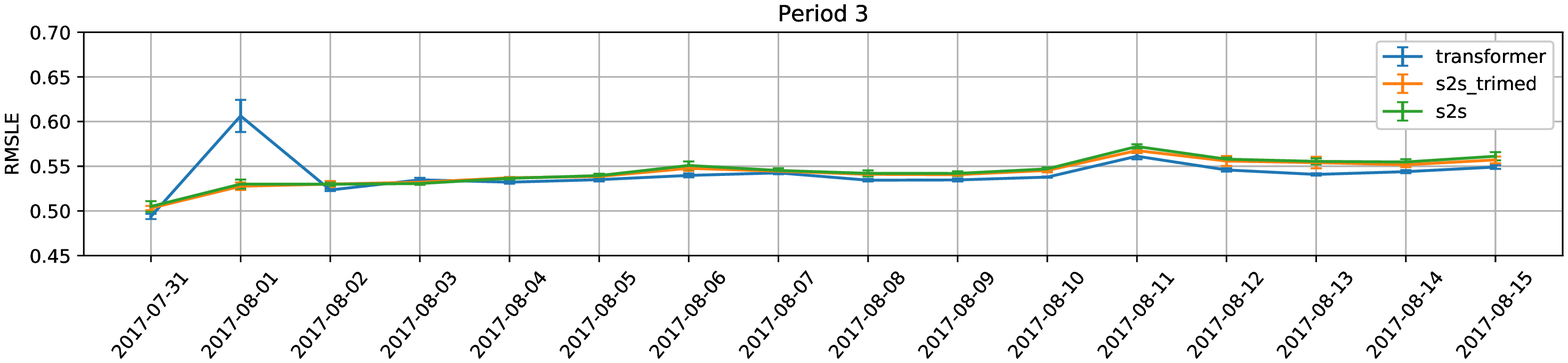}
	\caption{Daily \textit{RMSLE} for the three test periods (in chronological order, from top to bottom). The error bars shown in the figures represent the standard deviation of the three runs. Despite the error spikes in the 2nd day of the forecast, an ANOVA test shows non-significant differences between the average performance of the three models (with the following p-values: 0.0947, 0.1823 and 0.6181 for periods 1, 2 and 3, respectively). }
	\label{fig:dailyerror}
\end{figure}

From Figure \ref{fig:stores_items_performance}, we observe that the performance is very similar across models. A numerical comparison of the errors obtained for every model is presented in Table \ref{tab:results}. Additionally, a deeper daily analysis is provided in Figure \ref{fig:dailyerror}. From these figures, we conclude that the error is not distributed randomly across products, stores and time. 

	From Table \ref{tab:results}, we can conclude that the three models perform similarly. However, the daily figures show that the transformer error has more variability around the second and fourth day of forecast. This may be due to the fact that the model has been trained using \textit{teacher forcing} (\cite{williams1989, goyal2016}), and at inference time, an auto-regressive strategy has been used to compute the forecasted sales. This may cause distribution shifts that impact the quality of the forecast.  Besides, the \textit{seq2seq} models were much faster at training and inference time. This is due to the quadratic complexity dependence on the sequence length in the \textit{transformer} architecture; in \textit{seq2seq} it is linear. The simplest model (\textit{seq2seq trimmed}) was the fastest of the three alternatives, with no noticeable decrease in performance, either in the general picture or in the daily figures.

	\begin{table}[!h]
	\scriptsize
	\caption{Results of the models trained for three different time spans. All the models have been trained five times to reduce the effect of different random initialization. The errors are represented as mean $\pm$ standard deviation, across the five runs. We have highlighted the rows corresponding with the models that achieved the lowest RMSLE. At the bottom of the table we have included two benchmark metrics extracted from the previous works, although the authors do not clearly specify the period of time used to measure the results, they should be used only as a reference.}
		\label{tab:results}
		\centering
		\begin{tabular}{lllll}
			\hline
			Period    & Model                     & RMSLE                          & RMSWLE                         & MALE                           \\ \hline
			1         & \textbf{Seq2seq}          & $ \mathbf{0.5380 \pm 0.0016} $ & $ \mathbf{0.5376 \pm 0.0016 }$ & $  \mathbf{0.3450 \pm 0.0024}$ \\
			          & Seq2seq trimmed           & $ 0.5381 \pm 0.0008 $          & $ 0.5377 \pm 0.0008 $          & $ 0.3442 \pm 0.0008 $          \\
			          & Transformer               & $ 0.5439 \pm 0.0024 $          & $ 0.5436 \pm 0.0023 $          & $ 0.3386 \pm 0.001 $           \\
			          & Baseline: random          & $ 1.474 \pm 0.0003 $           & $ 1.4795 \pm 0.0003 $          & $ 1.0691 \pm 0.0002 $          \\
			          & Baseline: average         & $ 1.0422$                      & $ 1.05$                        & $ 0.8744$                      \\ \hline
			2         & Seq2seq                   & $ 0.5431 \pm 0.0014 $          & $ 0.5421 \pm 0.0013 $          & $ 0.3475 \pm 0.0012 $          \\
			          & \textbf{Seq2seq trimmed } & $ \mathbf{0.5413 \pm 0.0019} $ & $ 0.\mathbf{5403 \pm 0.0018 }$ & $ \mathbf{0.3444 \pm 0.0012} $ \\
			          & Transformer               & $ 0.5495 \pm 0.0021 $          & $ 0.5486 \pm 0.0021 $          & $ 0.3415 \pm 0.0012 $          \\
			          & Baseline: random          & $ 1.4649 \pm 0.0002 $          & $ 1.4702 \pm 0.0002 $          & $ 1.0577 \pm 0.0003 $          \\
			          & Baseline: average         & $ 1.0358$                      & $ 1.0433$                      & $ 0.8655$                      \\ \hline
			3         & Seq2seq                   & $ 0.544 \pm 0.0021 $           & $ 0.5431 \pm 0.0021 $          & $ 0.3502 \pm 0.0028 $          \\
			          & Seq2seq trimmed           & $ 0.5423 \pm 0.0015 $          & $ 0.5414 \pm 0.0016 $          & $ 0.3481 \pm 0.0017 $          \\
			          & \textbf{Transformer}      & $ \mathbf{0.5414 \pm 0.0015} $ & $ \mathbf{0.5407 \pm 0.0014} $ & $ \mathbf{0.3366 \pm 0.0012} $ \\
			          & Baseline: random          & $ 1.4555 \pm 0.0002 $          & $ 1.4606 \pm 0.0002 $          & $ 1.0517 \pm 0.0002 $          \\
			          & Baseline: average         & $ 1.029$                       & $ 1.0363 $                     & $ 0.8616$                      \\ \hline
			Benchmark & \cite{kechyn2018}         & -                              & 0.578                          & -                              \\
			Benchmark & \cite{Steves2018}         & -                              & 0.555                          & -                              \\ \hline
		\end{tabular}
	\end{table}

	\subsection{Ablation study}
	In this subsection we analyse the effect of two core pieces of our proposed architecture: the \textit{random max time step trick} and the length of the  input sequences in the \textit{Seq2seq trimmed} model.
	
	\subsubsection{Random max time step trick}
\begin{figure}[h!]
		\centering
		\includegraphics[width=0.7\linewidth]{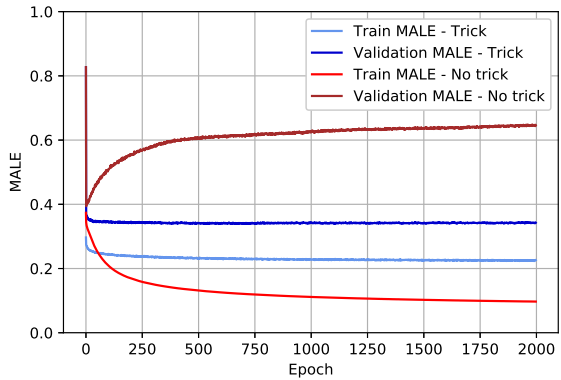}
		\caption{Training and validation MALE curves of the \textit{Seq2seq trimmed} when using the \textit{random max time step trick} and when disabling it.}
		\label{fig:traincurvesnotrick}
	\end{figure}

	We retrained the \textit{Seq2seq trimmed} model without the \textit{random max time step trick}. Table \ref{tab:results_notrick} summarizes the errors obtained at the best iteration of each model, averaged across five repeated runs. From the results, we conclude that when the \textit{random max time step trick} is used the model achieves significantly superior performance. This is due to the fact that randomizing the max time step helps the model to capture behaviors of the target signal at different times. Figure \ref{fig:traincurvesnotrick}, shows the training curves when the trick is used and when it is not used, suggesting that the trick may also act as a regularization technique, as the model is much less prone to overfitting when the trick is used. 
	
		\begin{table}[!h]
		\footnotesize
		\caption{Results of the models trained with and without the \textit{random max time step trick}. All the models have been trained five times to reduce the effect of different random initialization. The errors are represented as mean $\pm$ standard deviation, across the five runs. We have highlighted the rows corresponding with the models that achieved the lowest RMSLE}
		\label{tab:results_notrick}
		\centering
		\begin{tabular}{lllll}
			\hline
			Period & Trick/No trick             & RMSLE                 & RMSWLE                & MALE                  \\ \hline
			1 & Trick   & $ \mathbf{0.5381 \pm 0.0008} $ & $ \mathbf{0.5377 \pm 0.0008} $ & $ \mathbf{0.3442 \pm 0.0008} $ \\
			& No trick & $ 0.6077 \pm 0.0055 $  &  $ 0.6073 \pm 0.0054 $  &  $ 0.4037 \pm 0.0171 $ \\
			\hline
			2 & Trick  & $ \mathbf{0.5413 \pm 0.0019} $ & $ \mathbf{0.5403 \pm 0.0018} $ & $ \mathbf{0.3444 \pm 0.0012} $ \\
			& No trick & $ 0.5895 \pm 0.0042 $  &  $ 0.5886 \pm 0.0042 $  &  $ 0.3892 \pm 0.0216 $ \\
			\hline
			3 & Trick  & $ \mathbf{0.5423 \pm 0.0015} $ & $ \mathbf{0.5414 \pm 0.0016} $ & $ \mathbf{0.3481 \pm 0.0017} $ \\
			& No trick & $ 0.5929 \pm 0.0127 $  &  $ 0.5922 \pm 0.0125 $  &  $ 0.3938 \pm 0.0318 $ \\
			\hline
		\end{tabular}
	\end{table}

	\subsubsection{Input sequences length}
	As we showed in the Table \ref{tab:results}, the \textit{Seq2seq} model can be further simplified by trimming the length of the input sequences. In this subsection we study the minimum length of the input sequence without significant performance degradation. For that, we retrained the \textit{Seq2seq} model with different input sequence lengths to determine where is the optimum. The Table 		\ref{tab:results_ablation_length} shows the results of the model with the full sequences, and with sequences trimmed to 200, 75, 10, 1 and 0 time steps (where 0 time steps means not using any input sequence information at all, only static features). The results show that we can reduce the sequence lengths to up to 75 time steps without losing performance (and even slightly improving the generalization). Further reductions to 10 and 1 time steps start showing performance degradation. When not using input sequences (length=0) the performance degrades notably, as compared to using only one time step. We hypothesize that this happens because the model needs some reference level of number of sales per product to produce accurate forecasts. 
	
		\begin{table}[!h]
		\footnotesize
		\caption{Results of the \textit{Seq2seq} models trained with different input sequence lengths. All the models have been trained five times to reduce the effect of different random initialization. The errors are represented as mean $\pm$ standard deviation, across the five runs. We have highlighted the rows corresponding with the models that achieved the lowest RMSLE. An asterisk is added to the experiments where the metric is significantly different than the full sequence error (as per a two-tail T-test with $\alpha=0.05$)}
		\label{tab:results_ablation_length}
		\centering
		\begin{tabular}{lllll}
			\hline
			Period & Sequence length & RMSLE                           & RMSWLE                         & MALE                           \\ \hline
			1      & Full            & $ 0.5380 \pm 0.0016 $           & $ 0.5376 \pm 0.0016 $          & $  0.3450 \pm 0.0024$          \\
			       & 200             & $ 0.5381 \pm 0.0008 $           & $ 0.5377 \pm 0.0008 $          & $ 0.3442 \pm 0.0008 $          \\
			       & \textbf{75}     & $ \mathbf{0.5372 \pm 0.0016} $  & $ \mathbf{0.5369 \pm 0.0016} $ & $ \mathbf{0.3458 \pm 0.0034} $ \\
			       & 10              & $ 0.5452 \pm 0.0008 *$          & $ 0.5447 \pm 0.0009 *$         & $ 0.3478 \pm 0.0017 $          \\
			       & 1               & $ 0.5812 \pm 0.0042 *$          & $ 0.5807 \pm 0.0042 *$         & $ 0.3795 \pm 0.0029 *$          \\
			       & 0               & $ 0.8411 \pm 0.0016 *$          & $ 0.8453 \pm 0.0015 *$         & $ 0.5408 \pm 0.0028 *$          \\ \hline
			2      & Full            & $ 0.5431 \pm 0.0014 $           & $ 0.5421 \pm 0.0013 $          & $ 0.3475 \pm 0.0012 $          \\
			       & 200             & $ 0.5413 \pm 0.0019 $           & $ 0.5403 \pm 0.0018 $          & $ 0.3444 \pm 0.0012 *$          \\
			       & \textbf{75}     & $ \mathbf{0.5400 \pm 0.0010}* $ & $ \mathbf{0.5392 \pm 0.001}* $ & $ \mathbf{0.3458 \pm 0.0009} $ \\
			       & 10              & $ 0.5510 \pm 0.0049 *$          & $ 0.5501 \pm 0.0048 *$         & $ 0.3509 \pm 0.0038 $          \\
			       & 1               & $ 0.6162 \pm 0.0035 *$          & $ 0.6156 \pm 0.0037 *$         & $ 0.4011 \pm 0.0032 *$          \\
			       & 0               & $ 0.8426 \pm 0.0016 *$          & $ 0.8461 \pm 0.0015 *$         & $ 0.5388 \pm 0.0027 *$          \\ \hline
			3      & Full            & $ 0.5440 \pm 0.0021 $           & $ 0.5431 \pm 0.0021 $          & $ 0.3502 \pm 0.0028 $          \\
			       & 200             & $ 0.5423 \pm 0.0015 $           & $ 0.5414 \pm 0.0016 $          & $ 0.3481 \pm 0.001 $           \\
			       & \textbf{75}     & $ \mathbf{0.5418 \pm 0.0026} $  & $ \mathbf{0.5411 \pm 0.0025} $ & $ \mathbf{0.3499 \pm 0.0047} $ \\
			       & 10              & $ 0.5560 \pm 0.0051 *$          & $ 0.5548 \pm 0.0051 *$         & $ 0.3562 \pm 0.0049 $          \\
			       & 1               & $ 0.6360 \pm 0.0058 *$          & $ 0.6352 \pm 0.0062 *$         & $ 0.4163 \pm 0.0041 *$          \\
			       & 0               & $ 0.8387 \pm 0.0017 *$          & $ 0.8419 \pm 0.0016 *$         & $ 0.5373 \pm 0.0027 *$          \\ \hline
		\end{tabular}
	\end{table}
	
	\section{Conclusions} \label{sec:conclusions}
	
	Along this work, we have proposed a \textit{seq2seq} and a transformer architecture capable to solving the problem of sales forecasting for the \textit{Corporación Favorita} problem. We have also provided a trick (that we named \textit{random max time step trick}) that allowed to train the model to adapt to different time steps, not requiring to retrain the model every time a prediction is needed. We have empirically proved that it is possible to build a forecast for different products, at different points of sale and at different points in time using a single model. Our \textit{seq2seq trimmed} model achieved the best performance at the lowest theoretical computational cost. For that reason, we recommend its usage for this type of use cases.

	Deeper and more complex models must be tested further in order to try to improve performance by allowing more non-linear representations. In a real case, feature engineering may also be useful in order to help finding better representations. Finally, more sophisticated normalization methods for the target variable might be useful to to deal with different magnitudes and sparsity.
	
	\newpage

	\bibliography{mybib}

\begin{thebibliography}{31}
\expandafter\ifx\csname natexlab\endcsname\relax\def\natexlab#1{#1}\fi
\providecommand{\url}[1]{\texttt{#1}}
\providecommand{\href}[2]{#2}
\providecommand{\path}[1]{#1}
\providecommand{\DOIprefix}{doi:}
\providecommand{\ArXivprefix}{arXiv:}
\providecommand{\URLprefix}{URL: }
\providecommand{\Pubmedprefix}{pmid:}
\providecommand{\doi}[1]{\href{http://dx.doi.org/#1}{\path{#1}}}
\providecommand{\Pubmed}[1]{\href{pmid:#1}{\path{#1}}}
\providecommand{\bibinfo}[2]{#2}
\ifx\xfnm\relax \def\xfnm[#1]{\unskip,\space#1}\fi
\bibitem[{Badri et~al.(2017)Badri, Ghomi \& Hejazi}]{hossein2017}
\bibinfo{author}{Badri, H.}, \bibinfo{author}{Ghomi, S.}, \&
  \bibinfo{author}{Hejazi, T.~H.} (\bibinfo{year}{2017}).
\newblock \bibinfo{title}{Supply chain network design: A value-based approach}.
\newblock {\it \bibinfo{journal}{Transportation Research Part E}\/},  (pp.
  \bibinfo{pages}{1--17}). \DOIprefix\doi{10.1016/j.tre.2017.06.012}.
\bibitem[{Bahdanau et~al.(2015)Bahdanau, Cho \& Bengio}]{bahdanau2015}
\bibinfo{author}{Bahdanau, D.}, \bibinfo{author}{Cho, K.}, \&
  \bibinfo{author}{Bengio, Y.} (\bibinfo{year}{2015}).
\newblock \bibinfo{title}{Neural machine translation by jointly learning to
  align and translate}.
\newblock In {\it \bibinfo{booktitle}{3rd International Conference on Learning
  Representations}\/} ICLR'15.
\bibitem[{Bell(2000)}]{Bell2000}
\bibinfo{author}{Bell, P.~C.} (\bibinfo{year}{2000}).
\newblock \bibinfo{title}{Forecasting demand variation when there are
  stockouts}.
\newblock {\it \bibinfo{journal}{The Journal of the Operational Research
  Society}\/},  {\it \bibinfo{volume}{51}\/}, \bibinfo{pages}{358--363}.
  \URLprefix \url{http://www.jstor.org/stable/254094}.
\bibitem[{Calero \& Caro(2018)}]{Steves2018}
\bibinfo{author}{Calero, A. S.~M.}, \& \bibinfo{author}{Caro, J. M.~B.}
  (\bibinfo{year}{2018}).
\newblock {\it \bibinfo{title}{Corporación Favorita Grocery Sales
  Forecasting}\/}.
\newblock \bibinfo{type}{Master thesis}.
\newblock \URLprefix \url{https://dspace.unia.es/handle/10334/3921}.
\bibitem[{Corporación~Favorita(2018)}]{corporacionfavoritadataset2018}
\bibinfo{author}{Corporación~Favorita, K.} (\bibinfo{year}{2018}).
\newblock \bibinfo{title}{Corporación favorita grocery sales forecasting data
  set}.
\newblock \bibinfo{note}{Available in the following link:
  \url{https://www.kaggle.com/c/favorita-grocery-sales-forecasting/data}}.
\bibitem[{Cramer-Flood(2020)}]{emarketer2020}
\bibinfo{author}{Cramer-Flood, E.} (\bibinfo{year}{2020}).
\newblock \bibinfo{title}{Global ecommerce 2020: Ecommerce decelerates amid
  global retail contraction but remains a bright spot}.
\newblock {\it \bibinfo{journal}{E-marketer}\/}, .
\bibitem[{Curtin et~al.(2020)Curtin, Moseley, Ngo, Nguyen, Olteanu \&
  Schleich}]{Curtin2020}
\bibinfo{author}{Curtin, R.~R.}, \bibinfo{author}{Moseley, B.},
  \bibinfo{author}{Ngo, H.~Q.}, \bibinfo{author}{Nguyen, X.},
  \bibinfo{author}{Olteanu, D.}, \& \bibinfo{author}{Schleich, M.}
  (\bibinfo{year}{2020}).
\newblock \bibinfo{title}{Rk-means: Fast clustering for relational data}.
\newblock In \bibinfo{editor}{S.~Chiappa}, \& \bibinfo{editor}{R.~Calandra}
  (Eds.), {\it \bibinfo{booktitle}{The 23rd International Conference on
  Artificial Intelligence and Statistics, {AISTATS} 2020, 26-28 August 2020,
  Online [Palermo, Sicily, Italy]}\/} (pp. \bibinfo{pages}{2742--2752}).
\newblock \bibinfo{publisher}{Proceedings of Machine Learning Research} volume
  \bibinfo{volume}{108} of {\it \bibinfo{series}{Proceedings of Machine
  Learning Research}\/}.
\newblock \URLprefix \url{http://proceedings.mlr.press/v108/curtin20a.html}.
\bibitem[{Deep \& Salhi(2019)}]{Deep2019}
\bibinfo{author}{Deep, K.}, \& \bibinfo{author}{Salhi, M. J.~S.}
  (\bibinfo{year}{2019}).
\newblock {\it \bibinfo{title}{Logistics, Supply Chain and Financial Predictive
  Analytics}\/}.
\newblock \DOIprefix\doi{10.1007/978-981-13-0872-7}.
\bibitem[{Forslund \& Jonsson(2007)}]{forslund2007}
\bibinfo{author}{Forslund, H.}, \& \bibinfo{author}{Jonsson, P.}
  (\bibinfo{year}{2007}).
\newblock \bibinfo{title}{The impact of forecast quality on supply chain
  performance}.
\newblock {\it \bibinfo{journal}{International Journal of Operations \&
  Production Management}\/},  {\it \bibinfo{volume}{Vol.27}\/},
  \bibinfo{pages}{90--107}. \DOIprefix\doi{10.1108/01443570710714556}.
\bibitem[{Goodfellow et~al.(2016)Goodfellow, Bengio \&
  Courville}]{Goodfellow2016}
\bibinfo{author}{Goodfellow, I.}, \bibinfo{author}{Bengio, Y.}, \&
  \bibinfo{author}{Courville, A.} (\bibinfo{year}{2016}).
\newblock {\it \bibinfo{title}{Deep Learning}\/}.
\newblock \bibinfo{publisher}{The MIT Press}.
\bibitem[{Goyal et~al.(2016)Goyal, Lamb, Zhang, Zhang, Courville \&
  Bengio}]{goyal2016}
\bibinfo{author}{Goyal, A.}, \bibinfo{author}{Lamb, A.},
  \bibinfo{author}{Zhang, Y.}, \bibinfo{author}{Zhang, S.},
  \bibinfo{author}{Courville, A.}, \& \bibinfo{author}{Bengio, Y.}
  (\bibinfo{year}{2016}).
\newblock \bibinfo{title}{Professor forcing: A new algorithm for training
  recurrent networks}.
\newblock In {\it \bibinfo{booktitle}{Proceedings of the 30th International
  Conference on Neural Information Processing Systems}\/} NIPS'16 (p.
  \bibinfo{pages}{4608–4616}).
\bibitem[{Graves \& Willems(2008)}]{graves2008}
\bibinfo{author}{Graves, S.}, \& \bibinfo{author}{Willems, S.}
  (\bibinfo{year}{2008}).
\newblock \bibinfo{title}{Strategic inventory placement in supply chains:
  Nonstationary demand}.
\newblock {\it \bibinfo{journal}{Manufacturing \& Service Operations
  Management}\/},  {\it \bibinfo{volume}{Vol. 10}\/},
  \bibinfo{pages}{278--287}. \DOIprefix\doi{10.1287/msom.1070.0175}.
\bibitem[{Helmini et~al.(2019)Helmini, Jihan, Jayasinghe \&
  Perera}]{Helmini2019}
\bibinfo{author}{Helmini, S.}, \bibinfo{author}{Jihan, N.},
  \bibinfo{author}{Jayasinghe, M.}, \& \bibinfo{author}{Perera, S.}
  (\bibinfo{year}{2019}).
\newblock \bibinfo{title}{Sales forecasting using multivariate long short term
  memory network models}.
\newblock {\it \bibinfo{journal}{PeerJ Preprints}\/},  {\it
  \bibinfo{volume}{7}\/}, \bibinfo{pages}{e27712v1}. \URLprefix
  \url{https://doi.org/10.7287/peerj.preprints.27712v1}.
  \DOIprefix\doi{10.7287/peerj.preprints.27712v1}.
\bibitem[{Hyndman \& Athanasopoulos(2018)}]{Hyndman2018}
\bibinfo{author}{Hyndman, R.}, \& \bibinfo{author}{Athanasopoulos, G.}
  (\bibinfo{year}{2018}).
\newblock {\it \bibinfo{title}{Forecasting: Principles and Practice}\/}.
\newblock (\bibinfo{edition}{2nd} ed.).
\newblock \bibinfo{address}{Australia}: \bibinfo{publisher}{OTexts}.
\bibitem[{Kaipia(2009)}]{kaipia2009}
\bibinfo{author}{Kaipia, R.} (\bibinfo{year}{2009}).
\newblock \bibinfo{title}{Coordinating material and information flows with
  supply chain planning}.
\newblock {\it \bibinfo{journal}{International Journal of Logistics Management,
  The}\/},  {\it \bibinfo{volume}{Vol.20}\/}, \bibinfo{pages}{144--162}.
  \DOIprefix\doi{10.1108/09574090910954882}.
\bibitem[{Kechyn et~al.(2018)Kechyn, Yu, Zang \& Kechyn}]{kechyn2018}
\bibinfo{author}{Kechyn, G.}, \bibinfo{author}{Yu, L.}, \bibinfo{author}{Zang,
  Y.}, \& \bibinfo{author}{Kechyn, S.} (\bibinfo{year}{2018}).
\newblock \bibinfo{title}{Sales forecasting using wavenet within the framework
  of the kaggle competition}.
\newblock \href{http://arxiv.org/abs/1803.04037}{\tt arXiv:1803.04037}.
\bibitem[{Khamis et~al.(2020)Khamis, Ngo, Nguyen, Olteanu \&
  Schleich}]{Khamis2020}
\bibinfo{author}{Khamis, M.~A.}, \bibinfo{author}{Ngo, H.~Q.},
  \bibinfo{author}{Nguyen, X.}, \bibinfo{author}{Olteanu, D.}, \&
  \bibinfo{author}{Schleich, M.} (\bibinfo{year}{2020}).
\newblock \bibinfo{title}{Learning models over relational data using sparse
  tensors and functional dependencies}.
\newblock {\it \bibinfo{journal}{ACM Trans. Database Syst.}\/},  {\it
  \bibinfo{volume}{45}\/}. \URLprefix \url{https://doi.org/10.1145/3375661}.
  \DOIprefix\doi{10.1145/3375661}.
\bibitem[{Kilimci et~al.(2019)Kilimci, Akyuz, Uysal, Akyokus, Uysal, Bulbul \&
  Ekmis}]{Kilimci2019}
\bibinfo{author}{Kilimci, Z.~H.}, \bibinfo{author}{Akyuz, A.~O.},
  \bibinfo{author}{Uysal, M.}, \bibinfo{author}{Akyokus, S.},
  \bibinfo{author}{Uysal, M.~O.}, \bibinfo{author}{Bulbul, B.~A.}, \&
  \bibinfo{author}{Ekmis, M.~A.} (\bibinfo{year}{2019}).
\newblock \bibinfo{title}{An improved demand forecasting model using deep
  learning approach and proposed decision integration strategy for supply
  chain}.
\newblock {\it \bibinfo{journal}{Complexity}\/},  {\it
  \bibinfo{volume}{2019}\/}, \bibinfo{pages}{1--15}.
  \DOIprefix\doi{10.1155/2019/9067367}.
\bibitem[{Kuleshov et~al.(2018)Kuleshov, Fenner \& Ermon}]{Kuleshov2018}
\bibinfo{author}{Kuleshov, V.}, \bibinfo{author}{Fenner, N.}, \&
  \bibinfo{author}{Ermon, S.} (\bibinfo{year}{2018}).
\newblock \bibinfo{title}{Accurate uncertainties for deep learning using
  calibrated regression}.
\newblock (pp. \bibinfo{pages}{2796--2804}).
\bibitem[{Lim et~al.(2019)Lim, Arik, Loeff \& Pfister}]{Lim2019}
\bibinfo{author}{Lim, B.}, \bibinfo{author}{Arik, S.~{\"{O}}.},
  \bibinfo{author}{Loeff, N.}, \& \bibinfo{author}{Pfister, T.}
  (\bibinfo{year}{2019}).
\newblock \bibinfo{title}{Temporal fusion transformers for interpretable
  multi-horizon time series forecasting}.
\newblock {\it \bibinfo{journal}{CoRR}\/},  {\it
  \bibinfo{volume}{abs/1912.09363}\/}. \URLprefix
  \url{http://arxiv.org/abs/1912.09363}.
  \href{http://arxiv.org/abs/1912.09363}{\tt arXiv:1912.09363}.
\bibitem[{Lipsman(2019)}]{emarketer2019}
\bibinfo{author}{Lipsman, A.} (\bibinfo{year}{2019}).
\newblock \bibinfo{title}{Global ecommerce 2019: Ecommerce continues strong
  gains amid global economic uncertainty}.
\newblock {\it \bibinfo{journal}{E-marketer}\/}, .
\bibitem[{Malik et~al.(2019)Malik, Kuleshov, Song, Nemer, Seymour \&
  Ermon}]{Malik2019}
\bibinfo{author}{Malik, A.}, \bibinfo{author}{Kuleshov, V.},
  \bibinfo{author}{Song, J.}, \bibinfo{author}{Nemer, D.},
  \bibinfo{author}{Seymour, H.}, \& \bibinfo{author}{Ermon, S.}
  (\bibinfo{year}{2019}).
\newblock \bibinfo{title}{Calibrated model-based deep reinforcement learning}.
\newblock In \bibinfo{editor}{K.~Chaudhuri}, \&
  \bibinfo{editor}{R.~Salakhutdinov} (Eds.), {\it
  \bibinfo{booktitle}{Proceedings of the 36th International Conference on
  Machine Learning, {ICML} 2019, 9-15 June 2019, Long Beach, California,
  {USA}}\/} (pp. \bibinfo{pages}{4314--4323}).
\newblock \bibinfo{publisher}{Proceedings of Machine Learning Research}
  volume~\bibinfo{volume}{97} of {\it \bibinfo{series}{Proceedings of Machine
  Learning Research}\/}.
\newblock \URLprefix \url{http://proceedings.mlr.press/v97/malik19a.html}.
\bibitem[{van~den Oord et~al.(2016)van~den Oord, Dieleman, Zen, Simonyan,
  Vinyals, Graves, Kalchbrenner, Senior \& Kavukcuoglu}]{vanderoord2016}
\bibinfo{author}{van~den Oord, A.}, \bibinfo{author}{Dieleman, S.},
  \bibinfo{author}{Zen, H.}, \bibinfo{author}{Simonyan, K.},
  \bibinfo{author}{Vinyals, O.}, \bibinfo{author}{Graves, A.},
  \bibinfo{author}{Kalchbrenner, N.}, \bibinfo{author}{Senior, A.~W.}, \&
  \bibinfo{author}{Kavukcuoglu, K.} (\bibinfo{year}{2016}).
\newblock \bibinfo{title}{Wavenet: {A} generative model for raw audio}.
\newblock In {\it \bibinfo{booktitle}{The 9th {ISCA} Speech Synthesis Workshop,
  Sunnyvale, CA, USA, 13-15 September 2016}\/} (p. \bibinfo{pages}{125}).
\newblock \bibinfo{publisher}{{ISCA}}.
\newblock \URLprefix
  \url{http://www.isca-speech.org/archive/SSW\_2016/abstracts/ssw9\_DS-4\_van\_den\_Oord.html}.
\bibitem[{Schleich et~al.(2019)Schleich, Olteanu, Abo~Khamis, Ngo \&
  Nguyen}]{Schleich2019}
\bibinfo{author}{Schleich, M.}, \bibinfo{author}{Olteanu, D.},
  \bibinfo{author}{Abo~Khamis, M.}, \bibinfo{author}{Ngo, H.~Q.}, \&
  \bibinfo{author}{Nguyen, X.} (\bibinfo{year}{2019}).
\newblock \bibinfo{title}{A layered aggregate engine for analytics workloads}.
\newblock In {\it \bibinfo{booktitle}{Proceedings of the 2019 International
  Conference on Management of Data}\/} SIGMOD '19 (p.
  \bibinfo{pages}{1642–1659}).
\newblock \bibinfo{address}{New York, NY, USA}: \bibinfo{publisher}{Association
  for Computing Machinery}.
\newblock \URLprefix \url{https://doi.org/10.1145/3299869.3324961}.
  \DOIprefix\doi{10.1145/3299869.3324961}.
\bibitem[{Shaikhha et~al.(2020)Shaikhha, Schleich, Ghita \&
  Olteanu}]{Shaikhha2020}
\bibinfo{author}{Shaikhha, A.}, \bibinfo{author}{Schleich, M.},
  \bibinfo{author}{Ghita, A.}, \& \bibinfo{author}{Olteanu, D.}
  (\bibinfo{year}{2020}).
\newblock \bibinfo{title}{Multi-layer optimizations for end-to-end data
  analytics}.
\newblock In {\it \bibinfo{booktitle}{Proceedings of the 18th ACM/IEEE
  International Symposium on Code Generation and Optimization}\/} CGO 2020 (p.
  \bibinfo{pages}{145–157}).
\newblock \bibinfo{address}{New York, NY, USA}: \bibinfo{publisher}{Association
  for Computing Machinery}.
\newblock \URLprefix \url{https://doi.org/10.1145/3368826.3377923}.
  \DOIprefix\doi{10.1145/3368826.3377923}.
\bibitem[{Sutskever et~al.(2014)Sutskever, Vinyals \& Le}]{sutskever2014}
\bibinfo{author}{Sutskever, I.}, \bibinfo{author}{Vinyals, O.}, \&
  \bibinfo{author}{Le, Q.~V.} (\bibinfo{year}{2014}).
\newblock \bibinfo{title}{Sequence to sequence learning with neural networks}.
\newblock In {\it \bibinfo{booktitle}{Advances in Neural Information Processing
  Systems 27}\/} (pp. \bibinfo{pages}{3104--3112}).
\newblock \bibinfo{publisher}{Curran Associates, Inc.}
\bibitem[{Talupula(2018)}]{Talupula2018}
\bibinfo{author}{Talupula, A.} (\bibinfo{year}{2018}).
\newblock {\it \bibinfo{title}{Demand Forecasting Of Outbound Logistics Using
  Machine learning}\/}.
\newblock \bibinfo{type}{Master thesis}.
\newblock \URLprefix
  \url{https://www.diva-portal.org/smash/get/diva2:1367098/FULLTEXT02}.
\bibitem[{Uday~Kamath \& Whitaker(2019)}]{kamath2019}
\bibinfo{author}{Uday~Kamath, J.~L.}, \& \bibinfo{author}{Whitaker, J.}
  (\bibinfo{year}{2019}).
\newblock {\it \bibinfo{title}{Deep Learning for NLP and Speech
  Recognition}\/}.
\newblock (\bibinfo{edition}{1st} ed.).
\newblock \bibinfo{publisher}{Springer Publishing Company, Inc.}
\newblock \DOIprefix\doi{10.1007/978-3-030-14596-5}.
\bibitem[{Vaswani et~al.(2017)Vaswani, Shazeer, Parmar, Uszkoreit, Jones,
  Gomez, Kaiser \& Polosukhin}]{vaswani2017}
\bibinfo{author}{Vaswani, A.}, \bibinfo{author}{Shazeer, N.},
  \bibinfo{author}{Parmar, N.}, \bibinfo{author}{Uszkoreit, J.},
  \bibinfo{author}{Jones, L.}, \bibinfo{author}{Gomez, A.~N.},
  \bibinfo{author}{Kaiser, u.}, \& \bibinfo{author}{Polosukhin, I.}
  (\bibinfo{year}{2017}).
\newblock \bibinfo{title}{Attention is all you need}.
\newblock In {\it \bibinfo{booktitle}{Proceedings of the 31st International
  Conference on Neural Information Processing Systems}\/} NIPS'17 (p.
  \bibinfo{pages}{6000–6010}).
\bibitem[{{Wang} et~al.(2020){Wang}, {Cevik} \& {Bodur}}]{Wang2020}
\bibinfo{author}{{Wang}, J.}, \bibinfo{author}{{Cevik}, M.}, \&
  \bibinfo{author}{{Bodur}, M.} (\bibinfo{year}{2020}).
\newblock \bibinfo{title}{{On the Impact of Deep Learning-based Time-series
  Forecasts on Multistage Stochastic Programming Policies}}.
\newblock {\it \bibinfo{journal}{arXiv e-prints}\/},  (p.
  \bibinfo{pages}{arXiv:2009.00665}).
  \href{http://arxiv.org/abs/2009.00665}{\tt arXiv:2009.00665}.
\bibitem[{Williams \& Zipser(1989)}]{williams1989}
\bibinfo{author}{Williams, R.~J.}, \& \bibinfo{author}{Zipser, D.}
  (\bibinfo{year}{1989}).
\newblock \bibinfo{title}{A learning algorithm for continually running fully
  recurrent neural networks}.
\newblock {\it \bibinfo{journal}{Neural Computation}\/},  {\it
  \bibinfo{volume}{Vol. 1}\/}, \bibinfo{pages}{270--280}.

\end{thebibliography}
	

\end{document}